\documentclass{article}

\usepackage[preprint]{corl_2026} 

\usepackage{graphics}
\usepackage{graphicx}
\usepackage[table]{xcolor}
\usepackage{amssymb}
\usepackage{pifont}
\usepackage{tikz}
\usepackage{wrapfig}
\usepackage{makecell}
\usepackage{float}
\usepackage{amsmath}
\usepackage{subcaption}
\usepackage{caption}
\usepackage{algorithm}
\usepackage{algpseudocode}
\usepackage{titlesec}
\titlespacing*{\subsection}{0pt}{3pt}{2pt}
\titlespacing*{\section}{0pt}{5pt}{3pt}
\definecolor{myred}{HTML}{FF7F7F}
\definecolor{myblue}{HTML}{0040FF}
\definecolor{mygreen}{HTML}{09A309}
\definecolor{myyellow}{HTML}{D8D813}
\definecolor{myorange}{HTML}{FF8800}
\definecolor{mypurple}{HTML}{A600FF}
\definecolor{lightblue}{HTML}{D1ECF1}


\title{DIPOLE: Fusing Vision and Geometry for Robust Visuomotor Generalization}

  \author{
    Yikai Tang\textsuperscript{1*}, Haoran Geng\textsuperscript{1*$\dagger$}, Jindou Jia\textsuperscript{2}, Yuxuan Hu\textsuperscript{2},\\
    \bfseries Sheng Zang\textsuperscript{2}, Jianfei Yang\textsuperscript{2$\dagger$}, Pieter Abbeel\textsuperscript{1$\dagger$}, Jitendra
  Malik\textsuperscript{1$\dagger$}\\
    \textsuperscript{1}University of California, Berkeley\quad \textsuperscript{2}Nanyang Technological University\\
    * Equal Contribution, $\dagger$ Equal Advising
  }
  \AtBeginDocument{\hypersetup{
    pdftitle={DIPOLE: Fusing Vision and Geometry for Robust Visuomotor Generalization},
    pdfauthor={Yikai Tang, Haoran Geng, Jindou Jia, Yuxuan Hu, Sheng Zang, Jianfei Yang, Pieter Abbeel, Jitendra Malik},
    pdfsubject={},
    pdfkeywords={}
  }}
\begin{document}
\maketitle

\begin{figure}[h]
    \centering
    \includegraphics[width=\textwidth]{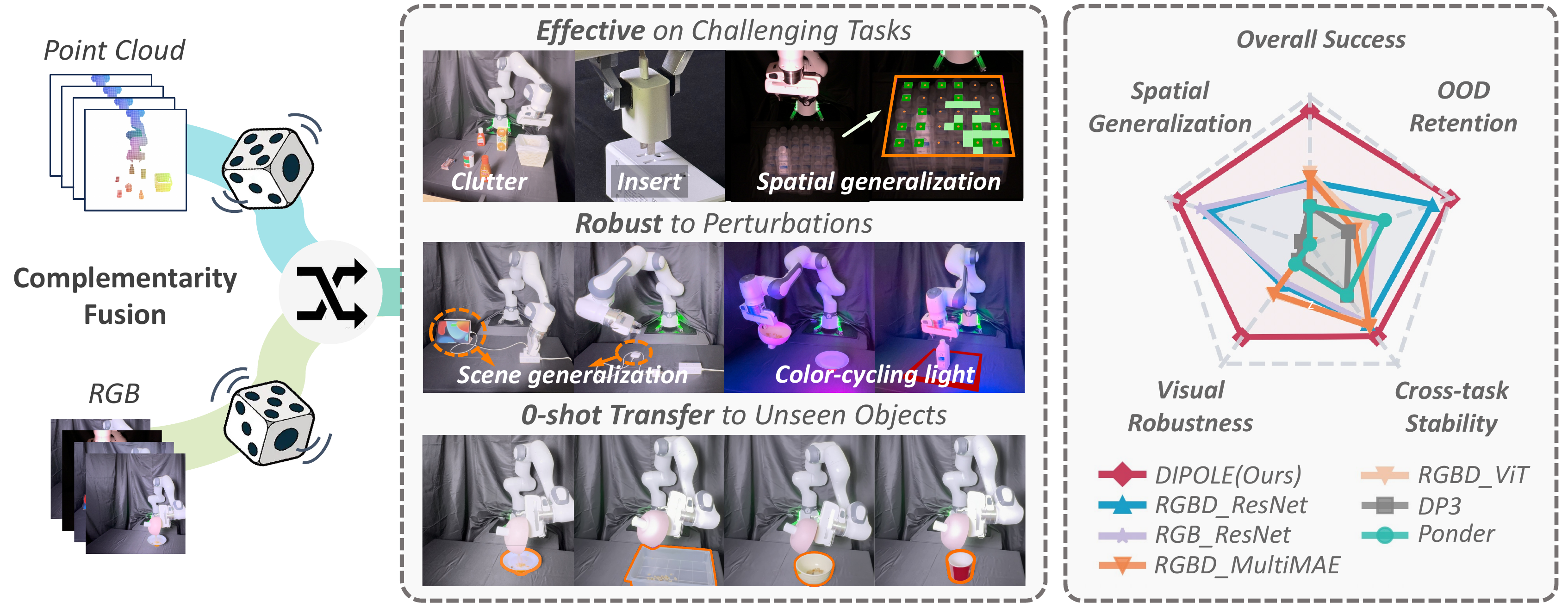}
    \caption{\small {\bf DIPOLE} is an imitation learning method that fuses 3D observations with 2D images through a \emph{complementarity-aware fusion module}, which uses modality-wise dropout to enforce balanced use of RGB and point cloud. This design yields substantial improvements in average performance, generalization, and robustness. DIPOLE is extensively evaluated in both simulation and the real world, covering a wide range of tasks and both visual and spatial uncertainties.}
    \label{fig:teaser}
\end{figure}

\begin{abstract}
Imitation learning has emerged as a crucial approach for acquiring visuomotor skills from demonstrations, where designing effective observation encoders is essential for policy generalization. However, existing methods tend to struggle once test-time conditions differ from the demonstrations, such as changes in lighting, texture, viewpoint, object placement, or object identity. To address this challenge, we propose \textbf{DI}ffusion \textbf{PO}licy with comp\textbf{L}ementarity \textbf{E}ncoders (\textbf{DIPOLE}), a visuomotor policy that learns to fuse complementary modalities through a training-time mechanism rather than a specialized fusion architecture. A modality-wise dropout masks one branch at each training step, encouraging each modality to remain individually informative. A lightweight cross-attention layer then exchanges complementary cues between the two. This design endows DIPOLE with five core strengths: stable high performance across diverse tasks, robustness to visual changes, spatial generalization at sub-centimeter precision, emergent capability beyond either modality, and zero-shot transfer to unseen objects. Across 18 simulated and 4 real-world tasks, DIPOLE outperforms six baselines by {39.1\%} on average, with gains of 41.5\% under unseen visual distractors and 15.2\% under randomized object placement.

\end{abstract}

\section{Introduction}
\label{sec:intro}

    Practical robot learning should produce policies that generalize beyond the laboratory into complex, unstructured environments. Recent methods reach strong performance on in-distribution evaluations~\cite{chi2024diffusionpolicyvisuomotorpolicy, ze20243ddiffusionpolicygeneralizable, brohan2023rt1roboticstransformerrealworld, zhao2023learningfinegrainedbimanualmanipulation}, yet break down once conditions shift in lighting, texture, viewpoint, object placement, or object identity~\cite{hu2024stemobgeneralizablevisualimitation, wang2024rise3dperceptionmakes, batra2025zeroshotvisualgeneralizationrobot,jia2026action}. The underlying reason is well understood, as robot demonstrations remain orders of magnitude scarcer than the text and image corpora driving modern foundation models~\cite{oxe2023}, leaving learned policies prone to overfitting. Extracting useful knowledge from limited demonstrations that remains robust across test-time distribution shifts has therefore become an increasingly critical research challenge.

\begin{wrapfigure}{r}{0.5\linewidth}
    \centering
    \vspace{-3mm}
    \includegraphics[width=\linewidth]{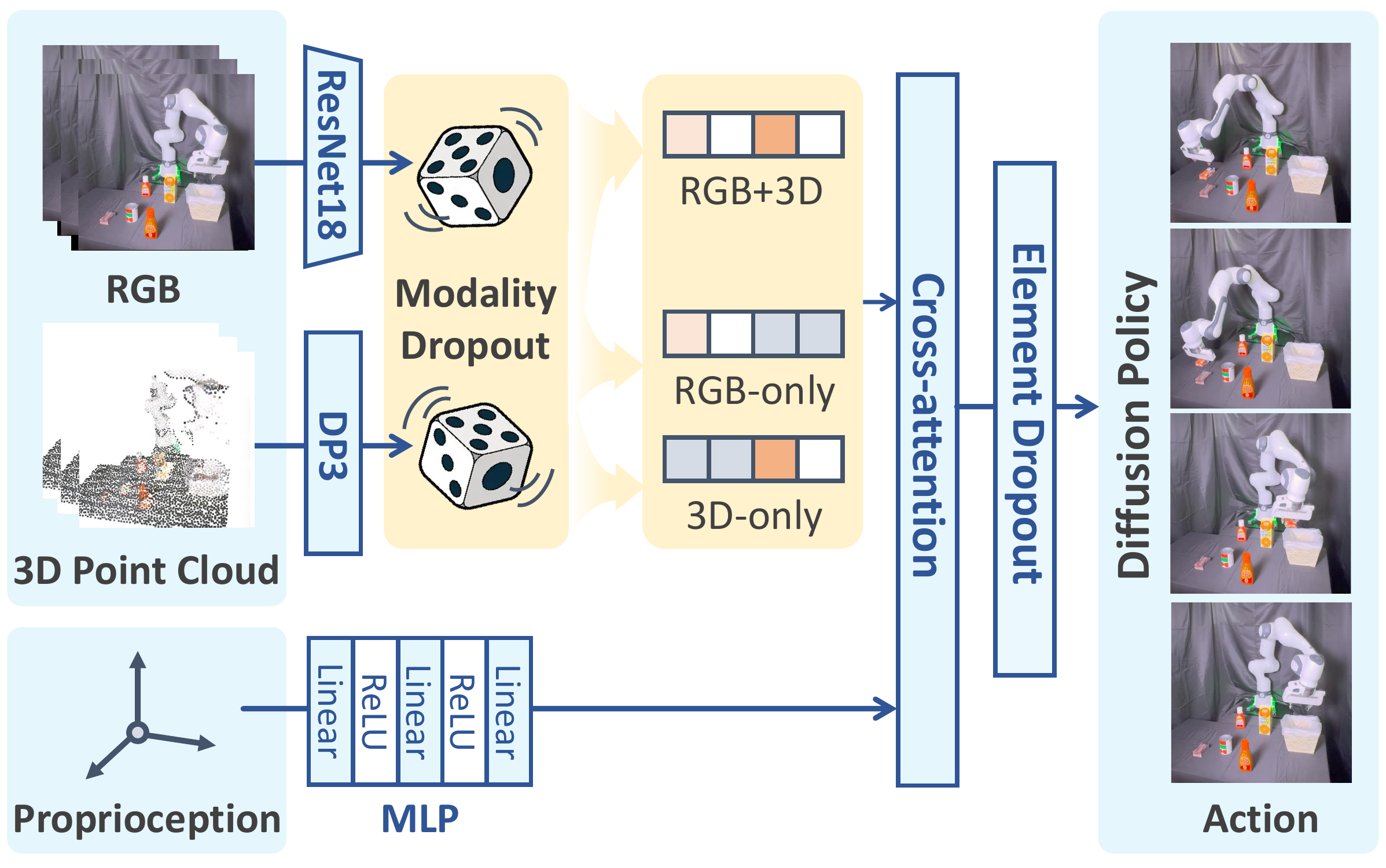}
    \vspace{-2mm}
    \setlength{\abovecaptionskip}{1pt}
    \setlength{\belowcaptionskip}{-2mm}
    \caption{
    {The architecture of DIPOLE.}
    }
    \label{fig:pipeline}
\end{wrapfigure}

   Existing methods aiming to address this gap mainly follow two lines. The first injects vision-language pretraining into robot policies to improve generalization, such as OpenVLA~\cite{kim2024openvla} and $\pi_{0.5}$~\cite{intelligence2025pi05}, which couple web-scale non-robotic data with growing pools of demonstrations. However, this paradigm pays a steep computational price, pushing pretraining costs into the range of thousands of GPU-hours~\cite{hu2024stemobgeneralizablevisualimitation}. More fundamentally, performance still degrades under distribution shifts in lighting, texture, viewpoint, object placement, or object identity, even after large-scale co-training~\cite{batra2025zeroshotvisualgeneralizationrobot}. The second line enriches observation modalities by employing point clouds~\cite{ze20243ddiffusionpolicygeneralizable}, depth~\cite{shridhar2022peract, goyal2023rvt}, or tactile signals~\cite{li2022shf} that expose geometry and contact-rich cues invisible to a 2D image. Although these modalities can surpass RGB in generalization under specific settings, each carries its own intrinsic limitations. While no single modality has emerged as uniformly reliable across diverse real-world environments~\cite{wang2024rise3dperceptionmakes, batra2025zeroshotvisualgeneralizationrobot}, a growing line of work fuses multiple modalities to obtain a richer representation~\cite{donat2025fusingpointcloudvisual, fu2025cordvipcorrespondencebasedvisuomotorpolicy}.

    These methods typically fuse RGB with point clouds or other geometric cues through fusion modules built on feature concatenation, late fusion, or cross-attention. When the two streams are relatively redundant, as for RGB and point cloud, this fusion usually brings little gain, since the multimodal policy collapses onto whichever stream is locally easier to fit and rarely breaks the ceiling of its dominant modality~\cite{xu2021fusionpaintingmultimodalfusionadaptive, wei2024robustmultimodallearningrepresentation, peng2022balancedmultimodal, wang2020what}. We refer to this phenomenon as \textit{modality collapse}. To address it, recent approaches either designate a primary modality and design a fusion mechanism that carefully injects information from the other modalities into the representation space of that primary modality~\cite{donat2025fusingpointcloudvisual}, or choose tactile signals that carry far less redundancy with RGB than depth, voxels, or point clouds, so as to avoid the impact of modality collapse~\cite{huang2024vitac}.

    However, humans do not behave this way. Studies of multisensory integration show that when people grasp, navigate, or insert objects, the brain combines visual, proprioceptive, and tactile signals simultaneously and continuously, weighting each cue by its momentary reliability rather than committing to any single dominant channel~\cite{ernst2002humans, stein2008multisensory}. This concurrent, reliability-weighted use of complementary senses is precisely what keeps everyday human behavior stable across lighting, texture, viewpoint, object placement, or object identity. This raises a natural question for robotics: can a visuomotor policy do the same, overcoming modality collapse by a similar mechanism so that all modalities act together rather than one dominating the rest?

    We answer this with \textbf{DI}ffusion \textbf{PO}licy with comp\textbf{L}ementarity \textbf{E}ncoders (\textbf{DIPOLE}), a visuomotor policy that predicts actions from RGB and geometric observations through a pair of complementary encoders. By design, each modality is preserved by how the policy is trained rather than by any specific fusion architecture. As shown in Fig.~\ref{fig:pipeline}, DIPOLE adopts a standard encoder for each modality, a ResNet-18 for RGB~\cite{he2015deepresiduallearningimage}, a DP3-style encoder for point clouds~\cite{ze20243ddiffusionpolicygeneralizable}, and a small MLP for proprioception. Before the features are fused and passed to the action head, we apply a simple yet effective mechanism called modality dropout. At each training step it masks either the RGB or the point-cloud feature, forcing the policy to reduce the training loss from each modality on its own rather than collapse onto either. The preserved features are then fused by a lightweight cross-attention layer and conditioned on a diffusion action head~\cite{chi2024diffusionpolicyvisuomotorpolicy}.

    Extensive experiments show that DIPOLE exhibits five key strengths. \textbf{Stable high performance across diverse tasks}. Across our 18-task simulation benchmark it attains a 65\% mean success rate, two to six times the 11\%--33\% baseline range, with the lowest per-task variation. \textbf{Robustness to visual changes}. Under material, table-position, and viewpoint shifts, it outperforms six baselines spanning RGB and RGBD encoders (ResNet~\cite{he2015deepresiduallearningimage}, ViT~\cite{dosovitskiy2020vit}, MultiMAE~\cite{bachmann2022multimae}) and point-cloud encoders (DP3~\cite{ze20243ddiffusionpolicygeneralizable}, PonderV2~\cite{zhu2023ponderv2}) by 41.5\% on average, with cross-randomization variance of 2.53\%, roughly 11$\times$ more stable than competing encoders. \textbf{Spatial generalization at sub-centimeter precision}. Grounded in a richer spatial representation, it raises the success rate by 15.2\% under randomized object placement and by 18.77\% on tasks requiring sub-centimeter precision. \textbf{Emergent capability beyond either modality}. It attains 72.3\% success on tasks where each unimodal encoder fails almost completely ($\leq$1\%). \textbf{Zero-shot transfer to unseen objects}. It generalizes to all six zero-shot transfer scenarios, spanning unseen objects, devices, and visual conditions. Aggregated across 18 simulated and 4 real-world tasks, DIPOLE outperforms all baselines by 39.1\% on average. Ablation studies indicate that these gains do not stem from any individual component, but from a fusion design that systematically enforces complementarity between modalities.

\section{Related work}

    {\bf \noindent Imitation learning.}
        Imitation learning acquires visuomotor control from demonstrations, avoiding costly online exploration~\cite{celemin2022interactiveimitationlearningrobotics,osa2018algorithmic}. Recent progress comes from two directions, novel policy formulations such as generative action modeling~\cite{chi2024diffusionpolicyvisuomotorpolicy} and richer observation modalities beyond RGB such as 3D point clouds~\cite{ze20243ddiffusionpolicygeneralizable}. Yet policies still degrade when the test scene differs from training, and common remedies such as domain randomization~\cite{tobin2017domain}, broader demonstration coverage~\cite{oxe2023}, and visual pretraining~\cite{nair2022r3m, majumdar2023vc1} narrow but rarely close this gap, since their effectiveness is bounded by what the observation encoder can represent~\cite{burns2023what}. We therefore strengthen the visual representation through complementarity-aware multimodal integration.

    {\bf \noindent Visual representation in robot learning.}
       Robots perceive through RGB, RGB-D, and point clouds, each trading off semantic detail against geometric invariance. RGB offers dense color and texture and is the default policy input~\cite{chi2024diffusionpolicyvisuomotorpolicy, brohan2023rt1roboticstransformerrealworld}, but its appearance basis is sensitive to lighting, texture, and viewpoint shifts~\cite{hu2024stemobgeneralizablevisualimitation}. Depth and RGB-D add appearance-invariant metric structure~\cite{shridhar2022peract, goyal2023rvt}, point clouds expose explicit 3D geometry that generalizes across viewpoint and placement but lack fine texture~\cite{ze20243ddiffusionpolicygeneralizable, wang2024rise3dperceptionmakes}, and tactile sensing captures contact forces invisible to cameras~\cite{li2022shf, huang2024vitac}. No single modality is uniformly reliable, which motivates fusing complementary modalities through explicit alignment and balanced use.

    {\bf \noindent Multimodal visual encoders.}
        Fusing RGB images with point clouds has become a mainstream approach for robotic perception and manipulation~\cite{donat2025fusingpointcloudvisual,fu2025cordvipcorrespondencebasedvisuomotorpolicy}. Multimodal visual encoders combine RGB’s rich semantics with 3D geometry’s metric structure~\cite{qi2017pointnetdeephierarchicalfeature}.  Despite these advances, challenges such as modality dominance \cite{komorowski2021minkloclidarmonocularimage,peng2022balancedmultimodal}, cross-sensor misalignment \cite{vora2020pointpaintingsequentialfusion3d}, and supervision imbalance persist \cite{xu2021fusionpaintingmultimodalfusionadaptive}. Motivated by these gaps, our method introduces a geometry-aware shared feature space and leverages attention mechanisms to achieve more balanced multimodal integration.

\section{Methodology}
    
      \subsection{Problem Formulation}
        \label{sec:formulation}

        We consider visuomotor imitation learning from a dataset of demonstrations $\mathcal{D} = \{(o_t, a_t)\}$, where each observation $o_t = (I_t, P_t, q_t)$ consists of an RGB image $I_t$, a point cloud $P_t$, and the proprioceptive state $q_t$. The goal is to learn a policy $\pi_\theta(a_t \mid o_t)$ using partial or all of the modalities that remains reliably effective under test-time shifts.

        We encode each modality independently with an off-the-shelf backbone: ResNet-18~\cite{he2015deepresiduallearningimage} for the RGB frame (512-d output), a DP3-style PointNet variant~\cite{ze20243ddiffusionpolicygeneralizable} for the point cloud (4{,}096 FPS-downsampled points, 128-d output), and an MLP for the joint state (64-d output). These encoder choices are standard, and can be replaced by any other encoder. DIPOLE's contribution lies in how the resulting per-modality features are fused.

    \subsection{\textbf{Modality Dropout for Complementarity-Aware Fusion}}
     \label{sec:integration}

        Building a policy that jointly and fully leverages all modalities, rather than collapsing onto a single dominant stream, runs into one well-documented obstacle, \emph{modality collapse}~\cite{wang2020what, peng2022balancedmultimodal, huang2022modalitycompetition}, in which the fused policy comes to rely on whichever modality is easier to fit while leaving the other underused. We diagnose this collapse formally and then introduce a \textit{simple yet efficient} mechanism to prevent it.

        \textbf{Setup.} Let $z_A = f_{\theta_A}(x_A)$ and $z_B = f_{\theta_B}(x_B)$ denote the per-modality features from Section~\ref{sec:formulation} (RGB and point cloud, respectively), and let $h$ be the fusion module producing $c = h(z_A, z_B)$. The fused representation conditions a policy $\pi(a \mid c)$ trained on demonstrations $\mathcal{D}$ with a loss $\mathcal{L}$.

        \textbf{Modality collapse issue.} Through the chain rule, the gradient reaching encoder $A$ factors as
        \begin{equation}
            \frac{\partial \mathcal{L}}{\partial \theta_A}
            \;=\;
            \frac{\partial \mathcal{L}}{\partial c}\,\frac{\partial c}{\partial z_A}\,\frac{\partial z_A}{\partial \theta_A},
            \label{eq:grad-A}
        \end{equation}
        with a symmetric expression for $\theta_B$. This decomposition is general: existing fusion designs, including concatenation, late linear fusion~\cite{peng2022balancedmultimodal}, and cross-attention, all correspond to specific choices of $h$ that admit the same failure mode. When modality $A$ alone is sufficient to drive $\mathcal{L}$ low, prior work has shown both empirically~\cite{wang2020what, peng2022balancedmultimodal} and theoretically~\cite{huang2022modalitycompetition} that the gradient magnitudes become severely imbalanced ($\|\partial \mathcal{L}/\partial \theta_A\| \gg \|\partial \mathcal{L}/\partial \theta_B\|$), reflecting a fusion module that effectively routes information through $z_A$ while suppressing the contribution of $z_B$ ($\|\partial c / \partial z_B\| \to 0$, which by Eq.~\eqref{eq:grad-A} starves encoder $B$ of training signal). 

    \textbf{Modality dropout.} We replace $c$ with
        \begin{equation}
            \tilde{c} = h(m_A \, z_A, \, m_B \, z_B),
            \qquad
            m_A, m_B \stackrel{\text{i.i.d.}}{\sim} \mathrm{Bernoulli}(1-p),
            \label{eq:masked-fusion}
        \end{equation}
        where $m_A, m_B \in \{0, 1\}$ are independent Bernoulli masks drawn at each training step. With probability $p$, modality $A$ is masked ($m_A = 0$), so the loss reduces to $\mathcal{L}\bigl(\pi(a \mid h(0, z_B))\bigr)$, which depends solely on $\theta_B$ and therefore produces a non-trivial gradient to encoder $B$ regardless of how easily $A$ could otherwise dominate. Symmetrically for $A$. Both encoders are thus guaranteed substantial gradient signal throughout training, breaking the dominance pattern admitted by static fusion.

        \textbf{Shared-latent attention.} Modality dropout prevents collapse, but a capable policy should also let each modality contribute in proportion to its instantaneous informativeness. We realize this with a shared-latent cross-attention layer. Because the two encoder outputs differ in dimensionality, 512-dimensional for RGB and 128-dimensional for point clouds, we first project them into a common 256-dimensional space through per-modality linear layers. We then apply a single bidirectional cross-attention layer with eight heads and one layer, which lets each modality query the other and reweight the exchanged information by learned attention scores.

    \subsection{Diffusion Action Head}
        \label{sec:decision}
        
        The fused representation $\mathbf{c}$ from Sec.~\ref{sec:integration} serves as the conditioning signal for the action head. DIPOLE adopts a diffusion policy~\cite{chi2024diffusionpolicyvisuomotorpolicy} for action selection. Starting from noisy action $a_T$, the policy iteratively denoises an action sequence conditioned on the fused context $\mathbf{c}$.
        
        \textbf{Training.} The action is noised for a random step $t$, i.e.,
            $
                \label{eq:Noising}
                a_t = \sqrt{1-\beta_t}\, a_{t-1} + \sqrt{\beta_t}\, \epsilon_t
            $
        where $\beta_t$ is a hyperparameter, and $\epsilon_t$ is Gaussian noise. In the diffusion step $t$, the network predicts the noise as
            $
                \hat{\boldsymbol{\epsilon}} = \epsilon_{\theta}\!\left(\mathbf{a}_t,\, t \,\middle|\, \mathbf{c}\right),
            $
        and is trained with the standard L2 objective
            $
                \mathcal{L}_{\text{diff}} = \mathbb{E}_{\mathbf{a}_0,\boldsymbol{\epsilon},t}\Big[ \big\| \boldsymbol{\epsilon} - \hat{\boldsymbol{\epsilon}} \big\|_2^2 \Big].
            $
        
        \textbf{Inference.} We initialize $\mathbf{a}_T\sim\mathcal{N}(\mathbf{0},\mathbf{I})$ and run the standard DDPM reverse process for $T$ steps, conditioning on the same $\mathbf{c}$ at every step, to produce the executed action chunk. Algorithm~\ref{alg:dipole} summarizes the full training procedure of DIPOLE.

        \vspace{-7pt}
        \begin{algorithm}[h]
            \caption{DIPOLE training}
            \label{alg:dipole}
            \begin{algorithmic}[1]
            \Require dataset $\mathcal{D} = \{(o_t, a_t)\}$ with $o_t = (I_t, P_t, q_t)$; dropout rate $p$; diffusion noise schedule $\{\beta_\tau\}_{\tau=1}^{T}$
            \Ensure trained encoders $f_{\theta_A}, f_{\theta_B}, f_{\theta_q}$, fusion module $h$, noise predictor $\epsilon_\theta$
            \Repeat
                \State Sample minibatch $\{(o_t, a_t)\} \sim \mathcal{D}$
                \State Encode per-modality features: $z_A \gets f_{\theta_A}(I_t)$, $z_B \gets f_{\theta_B}(P_t)$, $z_q \gets f_{\theta_q}(q_t)$
                \State Sample Bernoulli masks: $m_A, m_B \sim \mathrm{Bernoulli}(1-p)$ \Comment{modality dropout, Eq.~\eqref{eq:masked-fusion}}
                \State Fuse: $\tilde c \gets h(m_A\, z_A,\; m_B\, z_B,\; z_q)$ \Comment{shared-latent attention}
                \State Sample diffusion step $\tau \sim \mathcal{U}\{1, \dots, T\}$ and noise $\epsilon \sim \mathcal{N}(\mathbf{0}, \mathbf{I})$
                \State Noise the action: $a_\tau \gets \sqrt{1-\beta_\tau}\, a_{\tau-1} + \sqrt{\beta_\tau}\, \epsilon$
                \State Predict noise: $\hat{\epsilon} \gets \epsilon_\theta(a_\tau,\, \tau \mid \tilde c)$
                \State Update $\theta_A, \theta_B, \theta_q, h, \epsilon_\theta$ via gradient descent on $\mathcal{L}_{\mathrm{diff}} = \|\epsilon - \hat{\epsilon}\|_2^2$
            \Until{converged}
            \end{algorithmic}
        \end{algorithm}

\section{Experiments and analysis}
    To faithfully evaluate our method, we built a unified benchmark of 22 manipulation tasks (18 in simulation and 4 in the real world), tested against six baselines spanning RGB~\cite{he2015deepresiduallearningimage}, RGBD~\cite{dosovitskiy2020vit, bachmann2022multimae}, and point-cloud encoders~\cite{ze20243ddiffusionpolicygeneralizable, zhu2023ponderv2}. See Appendix~\ref{app:baselines} for full baseline details. Across these tasks DIPOLE outperforms all baselines by an average of 39.1\%, exhibiting five key features that arise from its complementarity-aware fusion design, as will be discussed later in this section.

\begin{figure}[h]
              \centering
              \includegraphics[width=0.8\linewidth]{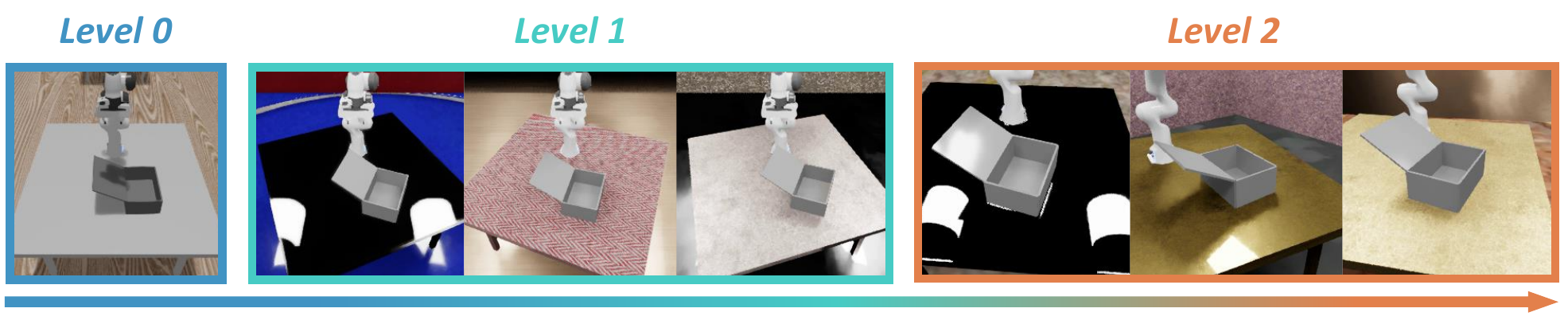}
              \caption{{\bf Environments under Different Randomization Levels.} The three brackets, named L0, L1, and L2, demonstrate example randomization in materials, positions, and viewpoints of the same task across different randomization levels.}
              \label{fig:vis_randomize}
          \end{figure}

    \textbf{Simulation benchmark.} The simulation suite draws tasks from LIBERO, RLBench, and ManiSkill~\cite{liu2023libero, james2020rlbench, tao2025maniskill3gpuparallelizedrobotics}, integrated through the RoboVerse platform~\cite{geng2025roboverse}, which exposes a unified API across the three sources. As shown in Fig.~\ref{fig:vis_randomize}, each task is evaluated under three levels of domain randomization (L0--L2), progressively adding material, table-position, and camera-viewpoint variations on top of the original task setup. As every method shares the same data pipeline, training schedule, and evaluation protocol, the resulting gaps solely reflect the policy design itself.

\begin{figure*}[!h]
              \centering
              \includegraphics[width=\linewidth]{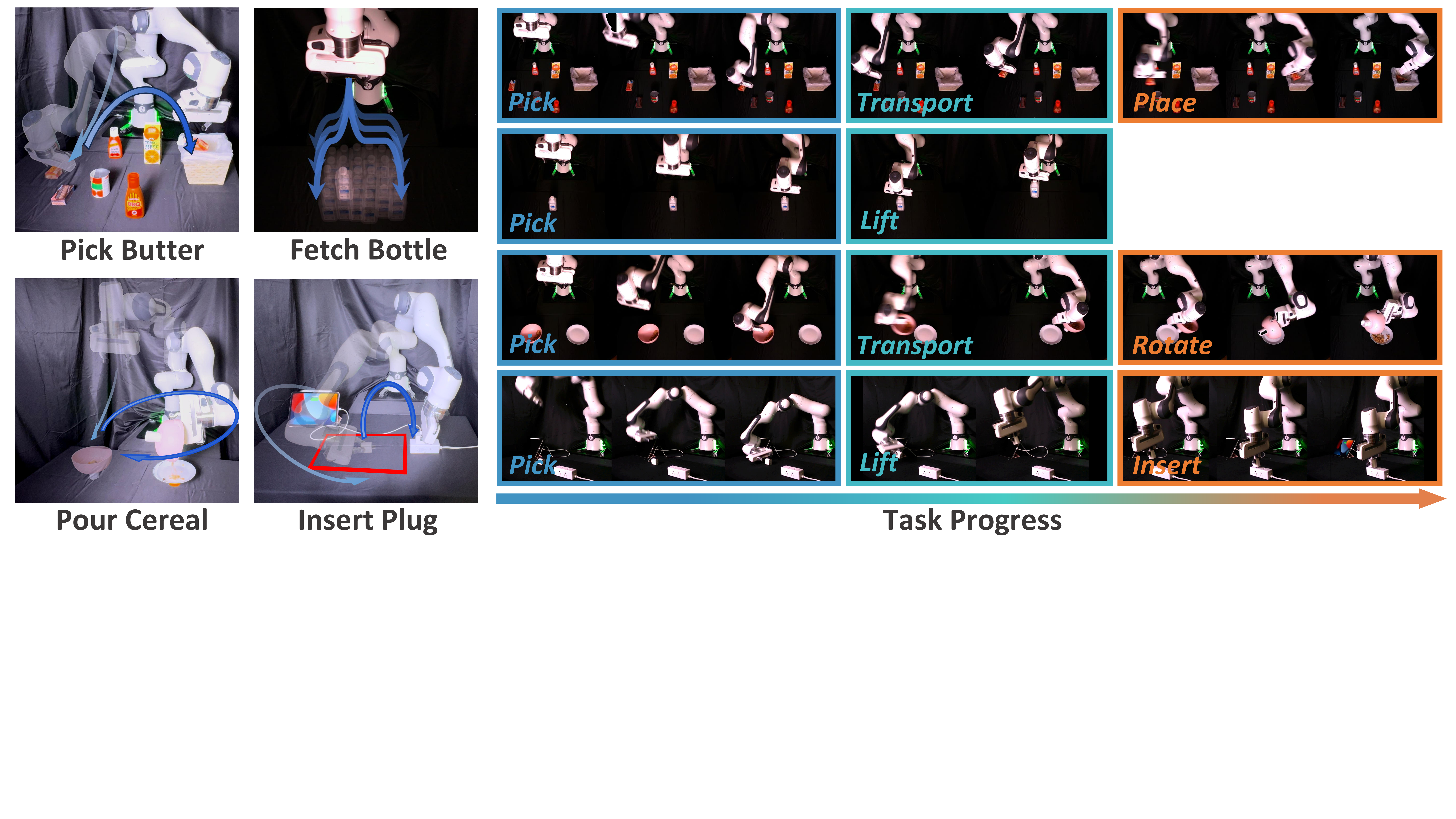}
              \caption{{\bf Real-world Benchmarks.} We deploy DIPOLE in the real world with a Franka Arm across four challenging tasks, including cluttered-scene manipulation (PickButter), 6-DoF cereal pouring (PourCereal), fine-grained handling under wide spatial generalization (FetchBottle), and contact-rich sub-centimeter insertion (InsertPlug).}
              \label{fig:realworld_demo}
          \end{figure*}

    \textbf{Real-world benchmark.} The real-world suite (Fig.~\ref{fig:realworld_demo}) covers four manipulation tasks spanning clutter, wide spatial randomization, 6-DoF orientation control, and contact-rich sub-centimeter insertion, with added test-time distractions in lighting, object placement, or object identity for evaluating generalization under unscripted variation.

     \textbf{Baseline choice.} We select six baselines that span the three mainstream observation modalities, RGB~\cite{he2015deepresiduallearningimage}, RGB-D~\cite{dosovitskiy2020vit, bachmann2022multimae}, and point clouds~\cite{ze20243ddiffusionpolicygeneralizable, zhu2023ponderv2}, with each modality represented by one or more strong encoders. Together they bound the best performance attainable from any single modality, so the comparison tests whether fusion can surpass the strongest unimodal encoder. Real-robot rollouts are far more time-consuming to run than simulated ones, so in the real world we compare against one representative encoder per modality, namely DP(RGB), DP(RGBD), and DP3.

        \begin{table}[h]
            \small
            \centering
            \begin{tabular}{l|cccc}
            \hline
            Task & \textbf{DIPOLE} & DP(RGB) & DP(RGBD) & DP3 \\
            \hline
            Pick Butter   & \cellcolor{lightblue}\textbf{100\%}   & 84\%    & 80\%    & 12\% \\ \hline
            Fetch Bottle  & \cellcolor{lightblue}\textbf{22.38\%} & 18\%    & 2.10\%  & 1.40\% \\ \hline
            Pour Cereal   & \cellcolor{lightblue}\textbf{100\%}   & 88\%    & 84\%    & 80.00\% \\ \hline
            Insert Plug   & \cellcolor{lightblue}\textbf{36\%}    & 25\%    & 16\%    & 0\% \\
            \hline
            \end{tabular}
            \vspace{6pt}
            \caption{{\bf Real-world policy performance across four tasks.} Highest score per task is highlighted.}
            \label{tab:realworld_policy}
        \end{table}
        \vspace{-10pt}
        
    \subsection{Balanced fusion yields emergent capabilities}
        \label{sec:f2}
        
        To examine whether DIPOLE's effectiveness extends beyond what either modality can achieve on its own, we test it on tasks where both unimodal baselines fail when trained alone. While both the RGB-only ResNet and the point-cloud-only DP3 baselines reach $\leq$1\% success when trained independently(Tab.~\ref{tab:emergence}), we discovered that DIPOLE recovers to 72.3\% mean success despite both constituent modalities being effectively at zero. The ablation in Tab.~\ref{tab:ablation} further isolates modality-wise dropout as the underlying mechanism, since removing this component causes the fused model to revert to the strongest unimodal baseline, no matter the choice of fusion methods.

        This shows that the emergent gain comes from how DIPOLE is trained rather than from the fusion architecture. More broadly, even a modality that is weak on a task can still supply complementary information that helps the policy represent the scene more faithfully, so fusion does not merely combine two strong streams but produces capabilities that neither modality alone can deliver.
        \begin{table}[h]
            \small
            \centering
            \begin{tabular}{l|cc|cc|cc}
            \hline
            & \multicolumn{2}{c|}{CBL0} & \multicolumn{2}{c|}{CBL1} & \multicolumn{2}{c}{CBL2} \\
            \cline{2-7}
            & IID & OOD & IID & OOD & IID & OOD \\
            \hline
            DIPOLE & \cellcolor{lightblue}\textbf{1.00} & \cellcolor{lightblue}\textbf{0.97}
                   & \cellcolor{lightblue}\textbf{0.94} & \cellcolor{lightblue}\textbf{0.95}
                   & \cellcolor{lightblue}\textbf{0.95} & \cellcolor{lightblue}\textbf{0.95} \\
            w/o Residual & 0.76 & 0.82 & 0.52 & 0.52 & 0.94 & 0.86 \\
            w/o Dropout  & 0.86 & 0.80 & 0.70 & 0.67 & \cellcolor{lightblue}\textbf{0.95} & 0.93 \\
            attention$\rightarrow$concat & 0.80 & 0.76 & 0.87 & 0.86 & 0.69 & 0.66 \\
            attention$\rightarrow$early fusion & 0.00 & 0.00 & 0.00 & 0.00 & 0.46 & 0.40 \\
            \hline
            DP (RGB-only)  & 0.81 & 0.52 & 0.40 & 0.20 & 0.42 & 0.32 \\
            DP3 (PC-only)  & 0.82 & 0.60 & 0.73 & 0.77 & 0.82 & 0.38 \\
            \hline
            \end{tabular}
            \caption{\textbf{Ablation study on CloseBox at three randomization levels.} DP = RGB-only Diffusion Policy; DP3 = point-cloud baseline.}
            \label{tab:ablation}
        \end{table}

    \vspace{-6pt}
    \subsection{DIPOLE obtains stable performance across diverse tasks}
        \label{sec:f1}
          To evaluate DIPOLE's effectiveness across a broader task suite covering diverse manipulation tasks, we benchmark it against six baselines on 18 simulated and 4 real-world tasks. DIPOLE attains a 65\% simulation mean (Tab.~\ref{tab:cross_task}), two to six times the 11\%--33\% baseline range, with the lowest per-task variation (Cov$_{\text{Task}}$ = 30.0\%, 1.5--4.4$\times$ smaller than baselines; Tab.~\ref{tab:stability}), and leads on every real-world task (Tab.~\ref{tab:realworld_policy}). High mean with low variance indicates a single generalist policy rather than one tuned to any task family.
        \begin{table}[!t]
            \centering
            \small
            \resizebox{\textwidth}{!}{%
            \begin{tabular}{l|ccccccc}
            \hline
            Task & \textbf{DIPOLE} & RGB\_resnet18 & RGBD\_resnet18 & RGBD\_ViT & RGBD\_MultiMAE & PointCloud\_DP3 & PointCloud\_PonderV2 \\
            \hline
            CloseBox    & \cellcolor{lightblue}\textbf{0.96} & 0.35 & 0.53 & 0.48 & 0.49 & 0.58 & 0.75 \\
            StackCube   & 0.03 & \cellcolor{lightblue}\textbf{0.1} & 0.06 & 0 & 0.01 & 0 & 0 \\
            AlphabetSoup& \cellcolor{lightblue}\textbf{0.79} & 0.22 & 0.24 & 0.61 & 0.48 & 0 & 0 \\
            Butter      & \cellcolor{lightblue}\textbf{0.88} & 0.72 & 0.62 & 0.68 & 0.69 & 0 & 0 \\
            OrangeJuice & \cellcolor{lightblue}\textbf{0.53} & 0.17 & 0.18 & 0.31 & 0.33 & 0.07 & 0.08 \\
            Tomato      & \cellcolor{lightblue}\textbf{0.72} & 0.08 & 0.11 & 0.13 & 0.14 & 0 & 0 \\
            \hline
            Average     & \cellcolor{lightblue}\textbf{0.65} & 0.27 & 0.29 & 0.32 & 0.33 & 0.11 & 0.14 \\
            \hline
            \end{tabular}}
            \caption{{\bf Cross-task performance comparison of simulation benchmark.} DIPOLE prevails over all others on almost all tasks, with an average absolute improvement of \textbf{39.1\%} and a smaller variance.\vspace{-10pt}}
            \label{tab:cross_task}
        \end{table}


    \subsection{DIPOLE generalizes stably to visual perturbations}
        \label{sec:f3}
         Real-world appearance drifts across materials, camera placement, and lighting, so we ask whether the geometric stream can carry DIPOLE when vision becomes unreliable. Visual shifts attack exactly the cues that the RGB stream depends on, which makes them a direct test of this question.

        Varying materials, table position, and camera viewpoint across three levels (L0 to L2, Fig.~\ref{fig:vis_randomize} and Tab.~\ref{tab:cross_randomization}), DIPOLE's success rate varies by only 2.53\%, roughly 11$\times$ smaller than the 28.5\% baseline average. At the most aggressive L2 it holds 64\% while baselines drop to 8\%--21\%. The invariance carries to real-world lighting, where DIPOLE passes the color-cycling Light-Strip zero-shot test (Fig.~\ref{fig:transfer}, Tab.~\ref{tab:zeroshot-crossobject}) while DP(RGB) and DP(RGBD) fail. The complementary geometric stream thus absorbs the visual perturbations that most degrade unimodal baselines.

    \subsection{Spatial generalization at sub-centimeter precision}
        \label{sec:f4}

         Geometry, in turn, is not always enough on its own. Sub-centimeter tasks demand accuracy that a downsampled point cloud is too coarse to resolve, even though it generalizes well across placement, so we ask the complementary question, whether the RGB stream's fine detail can carry precision where geometry falls short.

\begin{figure}[h]
          \centering
          \includegraphics[width=0.8\linewidth]{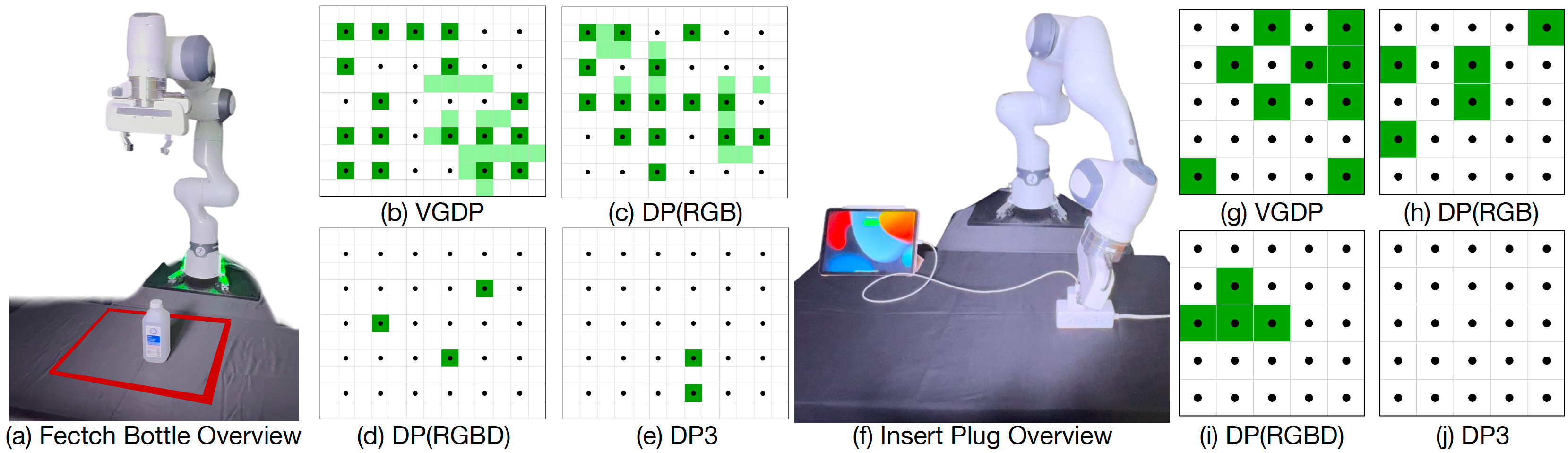}
          \caption{{\bf Real-world spatial randomization results.} \emph{FetchBottle}: policies trained on the 30 dotted grids and evaluated on all 143 positions; dark-green and light-green mark IID and OOD success, white marks failure. \emph{InsertPlug}: policies trained on all dotted grids and evaluated on 25 evenly distributed positions; dark-green marks success and white marks failure.}
          \label{fig:spatial_rand}
      \end{figure}

        On FetchBottle (Tab.~\ref{tab:realworld_policy} and Fig.~\ref{fig:spatial_rand}), trained on 30 grid positions and evaluated across the full 143-position workspace with a 2 cm bottle cap, DIPOLE reaches 22.4\% success against 1.4\%--18\% for the three baselines. On InsertPlug (Tab.~\ref{tab:realworld_policy} and Fig.~\ref{fig:spatial_rand}), which requires sub-centimeter vertical insertion across 25 grasp poses, DIPOLE leads at 36\% while RGB and RGBD degrade with placement shift and DP3 fails at the precision-critical grasp. The advantage carries to zero-shot deployment, where DIPOLE still succeeds after the iPad is moved to an unseen position (Fig.~\ref{fig:transfer}, Tab.~\ref{tab:zeroshot-crossobject}) while DP(RGBD) and DP3 fail. Complementarity thus lets DIPOLE inherit both RGB's fine-grained visual cues and the point cloud's placement-invariant structure, handling spatial generalization and sub-centimeter precision at once.

    \subsection{Zero-shot transfer across unseen objects}
        \label{sec:f5}

\begin{figure}[h]
            \centering
            \includegraphics[width=0.7\linewidth]{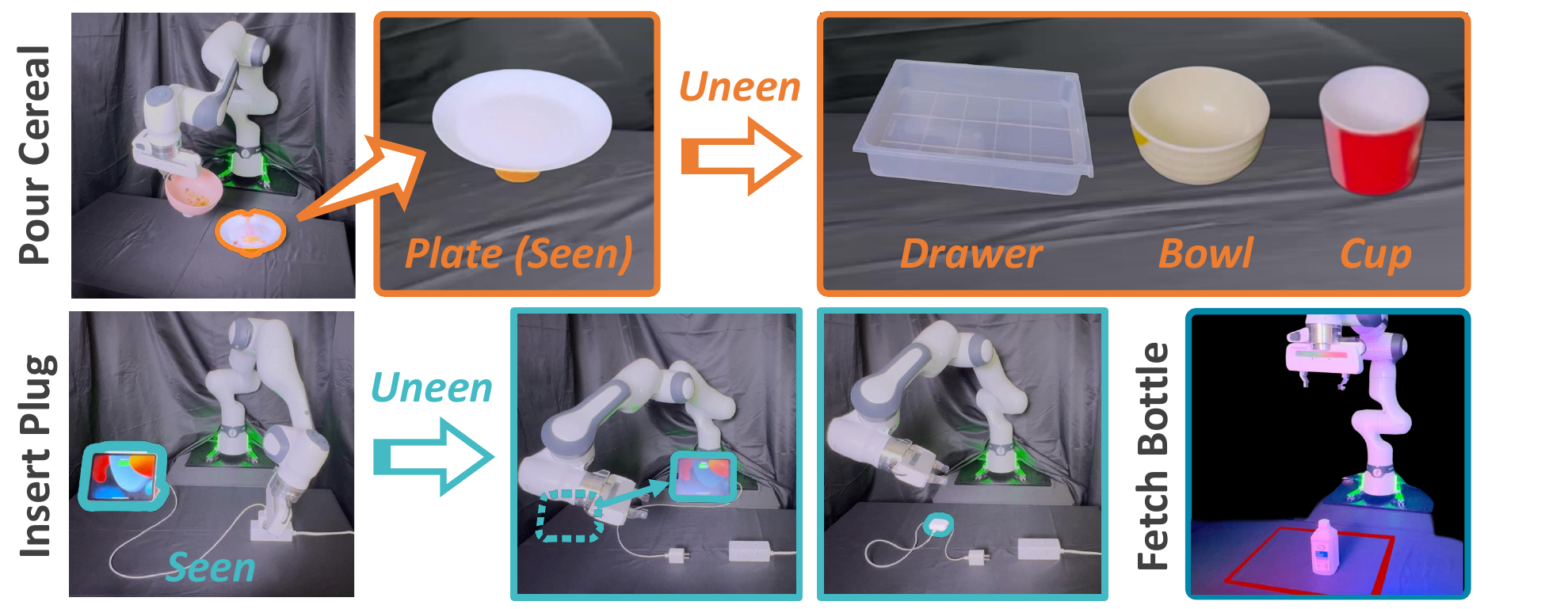}
            \caption{{\bf Zero-shot transfer settings.} (Top) In Pour-Cereal, the target plate is replaced with containers of varying sizes and shapes,
          including a drawer, a bowl, and a yogurt cup. (Bottom-Left) In Insert-Plug, the iPad is either moved to a novel position or replaced entirely with
          an AirPods charging case. (Bottom-Right) In Fetch-Bottle, the original lighting is swapped for a strong color-cycling illumination.}
            \label{fig:transfer}
        \end{figure}

        To evaluate DIPOLE's generalizability to fully unseen objects, we run four real-world zero-shot transfers in which the original target is replaced with novel instances. We replace the Pour-Cereal plate with a transparent drawer, a bowl, or a yogurt cup, and the Insert-Plug iPad with an AirPods charging case (Fig.~\ref{fig:transfer}, Tab.~\ref{tab:zeroshot-crossobject}). DIPOLE succeeds on all four transfers (4/4), while none of the unimodal baselines exceeds 2 of 4 successes. This pattern shows that DIPOLE generalizes by relying on whichever modality the novel instance still preserves in distribution, rather than requiring all streams to simultaneously remain intact. More broadly, robust generalization need not arise solely from scaling demonstration data. It can also emerge from observation representations that more faithfully encode the task-relevant structure of the scene.

        \begin{table}[h!]
            \small
            \renewcommand{\arraystretch}{0.95}
            \centering
            \setlength{\tabcolsep}{4pt}
            \begin{tabular}{c|cccccc}
                \hline
                    & Drawer & Bowl & Yogurt & Position & AirPods & Light Strip \\ \hline
                DIPOLE
                    & \textcolor{mygreen}{\checkmark} & \textcolor{mygreen}{\checkmark}
                    & \textcolor{mygreen}{\checkmark} & \textcolor{mygreen}{\checkmark}
                    & \textcolor{mygreen}{\checkmark} & \textcolor{mygreen}{\checkmark} \\ \hline
                DP(RGB)
                    & \textcolor{mygreen}{\checkmark} & \textcolor{mygreen}{\checkmark}
                    & \textcolor{red}{\ding{55}}      & \textcolor{mygreen}{\checkmark}
                    & \textcolor{red}{\ding{55}}      & \textcolor{red}{\ding{55}} \\ \hline
                DP(RGBD)
                    & \textcolor{mygreen}{\checkmark} & \textcolor{red}{\ding{55}}
                    & \textcolor{red}{\ding{55}}      & \textcolor{red}{\ding{55}}
                    & \textcolor{mygreen}{\checkmark} & \textcolor{red}{\ding{55}} \\ \hline
                DP3
                    & \textcolor{mygreen}{\checkmark} & \textcolor{mygreen}{\checkmark}
                    & \textcolor{red}{\ding{55}}      & \textcolor{red}{\ding{55}}
                    & \textcolor{red}{\ding{55}}      & \textcolor{mygreen}{\checkmark} \\ \hline
            \end{tabular}
            \caption{{\bf Zero-shot transfer across unseen objects, devices, and visual conditions.} DIPOLE succeeds on all six transfer scenarios, while unimodal baselines exhibit modality-specific weaknesses.}
            \label{tab:zeroshot-crossobject}
        \end{table}

\section{Conclusion}
     We present DIPOLE, a multimodal policy centered on a \emph{complementarity-aware fusion module} that enforces balanced use of RGB and 3D geometry so that neither modality dominates. Across 18 simulated and four real-world tasks featuring appearance shifts, spatial disturbances, and fine-grained control, DIPOLE outperforms strong 2D/3D/RGB-D baselines by an average of {39.1\%}, and attains {72.3\%} success on tasks where ResNet-only and DP3-only policies fail ($\leq$1\%). Ablations indicate that gains come from preserving rich RGB cues, enforcing balanced modality use, and adopting cross-attentive fusion over static combinations.

\section{Limitations \& Future Work}
     We have benchmarked DIPOLE mainly on tasks from LIBERO, RLBench, and ManiSkill, deploying only on a Franka Arm with Franka Gripper, using RoboVerse as the benchmarking platform. The performance of DIPOLE on higher-dimensional robot state space, e.g., Dexterous hands, on wider range of benchmarks, e.g. MetaWorld~\cite{yu2021metaworldbenchmarkevaluationmultitask} and on decision backbones beyond diffusion policy, e.g. Action Chunking~\cite{zhao2023learningfinegrainedbimanualmanipulation} is yet to be discovered. In future work, we will systematically explore the performance boundaries of DIPOLE by evaluating it across more diverse robotic platforms, task families, and control paradigms.


\clearpage
\acknowledgments{We would like to thank Yiran Wang and Feishi Wang for their helpful discussions and valuable feedback throughout this project. Pieter Abbeel holds concurrent appointments as a professor at the University of California, Berkeley and as an Amazon Scholar. This paper describes work performed at UC Berkeley and is not associated with Amazon.}


\bibliography{reference}
\clearpage

\appendix
\setcounter{table}{0}
\renewcommand{\thetable}{S\arabic{table}}

\section{Implementation Details}
\label{app:impl}

  We describe the full DIPOLE pipeline so that the policy can be reproduced
  end to end. The same architecture and training protocol are shared by all
  baselines, which differ only in the observation encoder (Appendix~\ref{app:baselines}).

  \paragraph{Observation preprocessing.}
  The environment is observed by a single-view RGB-D camera together with the
  robot joint state. The RGB frame is resized to $256{\times}256$ before encoding.
  The point cloud is obtained by back-projecting the depth map of the same
  RGB-D frame into a metric point cloud, cropping it to the task workspace,
  and downsampling it to $4{,}096$ points with farthest point sampling (FPS).
  The proprioceptive input is the vector of joint positions.

  \paragraph{Per-modality encoders.}
  RGB is encoded by a ResNet-18~\cite{he2015deepresiduallearningimage}, trained
  from scratch with BatchNorm replaced by GroupNorm, followed by global average
  pooling into a $512$-dimensional feature $z_A$. The point cloud is encoded by a
  DP3-style lightweight PointNet~\cite{ze20243ddiffusionpolicygeneralizable} into
  a $128$-dimensional feature $z_B$. The joint state is mapped by a small MLP into
  a $64$-dimensional feature $z_q$. These choices are standard and can be swapped
  for other encoders without changing the rest of the pipeline.

  \paragraph{Complementarity-aware fusion.}
  Before fusion we apply modality dropout. At each training step, the RGB or the
  point-cloud branch is independently masked with probability $p=0.2$
  (Eq.~\ref{eq:masked-fusion}), so that the loss is forced to flow through each
  encoder on its own and neither branch can dominate. The surviving features are
  linearly projected into a shared $256$-dimensional space and fused by a single
  bidirectional cross-attention layer, which lets each modality
  query the other and reweight the exchanged information. The cross-attention uses
  eight heads, the exchanged features are merged by summation, and a residual
  connection together with a learned modality encoding preserves the unimodal
  features before the fused representation $\tilde c$ is passed to the action head.

  \paragraph{Diffusion action head.}
  Conditioned on $\tilde c$, a diffusion policy~\cite{chi2024diffusionpolicyvisuomotorpolicy}
  predicts the action chunk with a conditional 1D U-Net. The policy conditions on a
  history of $3$ observation steps, predicts an action horizon of $8$, and executes
  the first $4$ predicted steps. Training minimizes the standard L2 noise-prediction
  loss (Sec.~\ref{sec:decision}) with a DDPM scheduler over $100$ timesteps and a
  cosine ($\mathrm{squaredcos\_cap\_v2}$) noise schedule; inference runs the full
  $100$-step DDPM reverse process from Gaussian noise, conditioning on the same
  $\tilde c$ at every denoising step. Algorithm~\ref{alg:dipole} gives the full
  training loop.

  \begin{table}[h]
    \small
    \centering
    \caption{Architecture and training hyperparameters of DIPOLE, taken from the
    released training configuration.}
    \label{tab:hyperparams}
    \begin{tabular}{l|l}
      \hline
      Component & Setting \\
      \hline
      \multicolumn{2}{l}{\emph{Observation and action}} \\
      RGB input resolution            & $256{\times}256$ \\
      Point-cloud points (FPS)        & $4{,}096$ \\
      Observation history $n_{\text{obs}}$ & $3$ \\
      Prediction horizon              & $8$ \\
      Executed action steps           & $4$ \\
      Action dimension                & $9$ (EE pose, quaternion, gripper) \\
      \hline
      \multicolumn{2}{l}{\emph{Encoders and fusion}} \\
      RGB encoder / output dim        & ResNet-18 (scratch, GroupNorm) / $512$ \\
      Point-cloud encoder / output dim& DP3 PointNet (LayerNorm) / $128$ \\
      Proprioception encoder / output dim & MLP / $64$ \\
      Shared fusion dim               & $256$ \\
      Cross-attention                 & bidirectional, $8$ heads, single layer \\
      Fusion merge / residual         & sum / yes (with modality encoding) \\
      Modality dropout rate $p$       & $0.2$ \\
      \hline
      \multicolumn{2}{l}{\emph{Diffusion head}} \\
      U-Net down dims                 & $[256, 512, 1024]$ \\
      Diffusion step embed dim        & $128$ \\
      Kernel size / groups            & $5$ / $8$ \\
      Noise scheduler                 & DDPM, $\mathrm{squaredcos\_cap\_v2}$ \\
      $\beta$ range                   & $[1{\times}10^{-4},\, 2{\times}10^{-2}]$ \\
      Train / inference timesteps     & $100$ / $100$ \\
      Prediction type                 & $\epsilon$ (noise) \\
      \hline
      \multicolumn{2}{l}{\emph{Optimization}} \\
      Optimizer                       & AdamW, $\beta=(0.95, 0.999)$, wd $1{\times}10^{-6}$ \\
      Learning rate / schedule        & $1{\times}10^{-4}$ / cosine, $500$ warmup steps \\
      Batch size                      & $128$ \\
      EMA                             & power $0.75$, max $0.9999$ \\
      Training (simulation)           & $1000$ epochs $\times\, 250$ steps \\
      Training (real world)           & $3000$ epochs \\
      Random seed                     & $42$ \\
      \hline
    \end{tabular}
  \end{table}

\section{Baseline Encoders}
\label{app:baselines}

  All baselines share the same diffusion-policy action head~\cite{chi2024diffusionpolicyvisuomotorpolicy}
  and proprioception MLP as DIPOLE, and differ only in the observation encoder.
  This design isolates the effect of the visual representation on policy
  performance. We compare against six encoders, grouped into RGB-only, RGB-D,
  and point-cloud families.

  \paragraph{RGB (ResNet-18).} The canonical 2D Diffusion Policy, which encodes only the RGB image with a ResNet-18 backbone~\cite{he2015deepresiduallearningimage}. It serves as the appearance-only reference and carries no explicit geometric input.

  \paragraph{RGB-D (ResNet-18).} The same ResNet-18 backbone applied to a four-channel RGB-D image, where the metric depth map is concatenated to the RGB channels. This is the simplest way to inject geometry into a 2D encoder.

  \paragraph{RGB-D (ViT).} A Vision Transformer~\cite{dosovitskiy2020vit} that tokenizes the RGB-D image into patches. It tests whether a stronger global-attention backbone, rather than a convolutional one, better exploits the added depth channel.

  \paragraph{RGB-D (MultiMAE).} A multi-modal masked autoencoder~\cite{bachmann2022multimae} that pretrains a shared ViT encoder across RGB and depth through cross-modal masked reconstruction. It represents the pretrained multi-modal representation-learning line of work, as opposed to training the fusion encoder from scratch.

  \paragraph{Point cloud (DP3).} 3D Diffusion Policy~\cite{ze20243ddiffusionpolicygeneralizable}, which encodes the back-projected point cloud with a lightweight PointNet-style network after farthest-point-sampling downsampling. It is the geometry-only counterpart to the RGB baseline.

  \paragraph{Point cloud (PonderV2).} A point-cloud encoder built on a SparseUNet backbone and pretrained with a differentiable rendering objective~\cite{zhu2023ponderv2}. It tests whether a heavier, pretrained 3D backbone improves over the lightweight DP3 encoder.

  \begin{table}[h]
    \small
    \centering
    \caption{Summary of baseline observation encoders. All variants share the same diffusion action head and proprioception encoder, and differ only in the observation modality and backbone.}
    \label{tab:baseline-summary}
    \begin{tabular}{l|lll}
        \hline
        Table name & Backbone & Input modality & Reference \\
        \hline
        RGB\_resnet18      & ResNet-18      & RGB    & \cite{he2015deepresiduallearningimage} \\
        RGBD\_resnet18     & ResNet-18      & RGB-D  & \cite{he2015deepresiduallearningimage} \\
        RGBD\_ViT          & ViT            & RGB-D  & \cite{dosovitskiy2020vit} \\
        RGBD\_MultiMAE     & MultiMAE (ViT) & RGB-D  & \cite{bachmann2022multimae} \\
        PointCloud\_DP3    & PointNet (DP3) & Point cloud & \cite{ze20243ddiffusionpolicygeneralizable} \\
        PointCloud\_PonderV2 & PonderV2 (SparseUNet) & Point cloud & \cite{zhu2023ponderv2} \\
        \hline
    \end{tabular}
  \end{table}

\section{Simulation Benchmark Details}
\label{app:sim}

  \paragraph{Platform and protocol.}
  All simulation experiments are run on the RoboVerse platform~\cite{geng2025roboverse},
  which integrates tasks from LIBERO~\cite{liu2023libero}, RLBench~\cite{james2020rlbench},
  and ManiSkill~\cite{tao2025maniskill3gpuparallelizedrobotics} behind a single
  API with a consistent simulator and configuration, so that all encoders are
  compared under identical data, training, and evaluation. Each task is trained
  with $100$ demonstrations and evaluated over $200$ trials, split evenly between
  in-distribution (IID) and out-of-distribution (OOD) scenes. The benchmark
  consists of six base manipulation tasks, CloseBox (RLBench), StackCube
  (ManiSkill), and AlphabetSoup, Butter, OrangeJuice, and Tomato (LIBERO),
  reported in main-text Table~\ref{tab:cross_task}. Combined with the three
  randomization levels of Table~\ref{tab:randomization-levels}, these give the
  $18$ simulated task settings reported in the main text.

  \paragraph{Randomization levels.}
  RoboVerse exposes controlled randomization of lighting, textures and materials,
  specular reflectance, and camera pose. We define three increasing levels
  (Table~\ref{tab:randomization-levels}). L0 is the original task with no
  randomization. L1 randomizes the wall, table, and object materials and rotates
  the table to an arbitrary orientation. L2 additionally randomizes the camera
  position by sampling from a predefined pool. Because every method sees the same
  randomization protocol, the resulting performance gaps reflect the policy design
  alone.

  \begin{table}[h]
    \small
    \centering
    \caption{Randomization factors at each level. L1 randomizes wall, table, and
    object materials and rotates the table; L2 further randomizes the camera
    position from a predefined pool.}
    \label{tab:randomization-levels}
    \begin{tabular}{l|ccc}
      \hline
      Level & Materials & Table position & Camera position \\
      \hline
      L0 & \textcolor{red}{\ding{55}}      & \textcolor{red}{\ding{55}}      & \textcolor{red}{\ding{55}} \\
      L1 & \textcolor{mygreen}{\checkmark} & \textcolor{mygreen}{\checkmark} & \textcolor{red}{\ding{55}} \\
      L2 & \textcolor{mygreen}{\checkmark} & \textcolor{mygreen}{\checkmark} & \textcolor{mygreen}{\checkmark} \\
      \hline
    \end{tabular}
  \end{table}

  \paragraph{IID / OOD split.}
  Within each task the $200$ evaluation trials are divided equally into IID scenes,
  drawn from the same randomization distribution seen during training, and OOD
  scenes, drawn from held-out values of the randomized factors. Reporting both
  lets us separate raw competence from generalization. The per-level success
  rates averaged over all tasks are given in Table~\ref{tab:cross_randomization},
  and the relative dispersion across tasks and levels in Table~\ref{tab:stability}.

\section{Real-World Benchmark Details}
\label{app:real}

  \paragraph{Hardware and data collection.}
  Real-world experiments use a Franka Emika Panda arm with a Franka gripper.
  Expert demonstrations are collected by human teleoperation with
  GELLO~\cite{wu2024gello}. Each policy is trained for $3000$ epochs on $100$
  demonstrations and evaluated over $25$ trials unless otherwise stated.
  Because real-robot rollouts are far more time-consuming than simulation, the
  real-world comparison keeps one representative encoder per modality, namely
  DP(RGB) (ResNet-18 on RGB), DP(RGBD) (ResNet-18 on RGB-D), DP3 (point cloud),
  and DIPOLE (multimodal).

  \paragraph{Tasks.}
  The four real-world tasks (Fig.~\ref{fig:realworld_demo}) are designed to probe
  distinct skills.

  \paragraph{PickButter.} Seven everyday objects of diverse sizes, shapes, and
  colors are placed on the desk, and the policy must pick the butter out of the
  cluttered scene and place it in the basket. The setting follows the
  corresponding simulation task so that the Sim2Real gap can be measured directly.

  \paragraph{FetchBottle.} A bottle is placed within a $39\,\text{cm}\times33\,\text{cm}$
  rectangle that covers most of the reachable workspace, and the policy must pick
  it up. The task demands precise control because the bottle cap is only
  $2\,\text{cm}$ in diameter. Training uses $30$ grid positions and evaluation
  spans the full $143$-position workspace (Fig.~\ref{fig:spatial_rand}).

  \paragraph{PourCereal.} A bowl of cereal and a plate are placed on the desk, and
  the policy must pick up the bowl and pour the cereal onto the plate. The
  difficulty lies in continuous control of 3D orientation, where small angular
  errors cause spillage.

  \paragraph{InsertPlug.} A tablet charging plug rests within a
  $20\,\text{cm}\times20\,\text{cm}$ area with a power strip on the opposite side.
  The policy must grasp the plug, lift it, move it above the power strip, and
  insert it into the socket so that the tablet begins charging. Insertion requires
  near-vertical alignment with enough force to overcome socket friction, and
  lateral deviation triggers a safety stop, so the task is sensitive to
  sub-centimeter error. Each trial starts from a different plug-socket
  orientation, and evaluation uses $25$ positions spread evenly across the
  workspace (Fig.~\ref{fig:spatial_rand}).

  \paragraph{Zero-shot transfer.}
  The zero-shot study (Sec.~\ref{sec:f5}, Tab.~\ref{tab:zeroshot-crossobject})
  replaces the PourCereal plate with a semi-transparent drawer, a bowl, or a
  yogurt cup, and replaces the InsertPlug iPad with an AirPods charging case or
  moves it to an unseen position. FetchBottle is additionally tested under strong
  color-cycling illumination. These six settings span semantic, geometric,
  visual, and spatial distribution shift.

\section{Ablation Variant Details}
\label{app:ablation}

  Main-text Table~\ref{tab:ablation} ablates the fusion design on CloseBox across
  the three randomization levels. We describe each variant here.

  \paragraph{w/o Dropout.} Removes modality dropout while keeping the
  cross-attention layer. Without the dropout constraint the fused model tracks the
  stronger unimodal encoder at each level, which shows that cross-attention alone
  does not prevent collapse.

  \paragraph{w/o Residual.} Removes the residual connection around the
  cross-attention layer, which reduces the stability of the fused representation.

  \paragraph{attention $\rightarrow$ concat.} Replaces cross-attention with feature
  concatenation under the same encoders. Passive late fusion struggles to exploit
  complementary structure and performs between the RGB-only and point-cloud-only
  baselines.

  \paragraph{attention $\rightarrow$ early fusion.} Replaces cross-attention with
  early fusion. This variant collapses almost entirely at L0 and L1 and recovers
  only slightly under heavy viewpoint variation, mirroring the point-cloud-only
  baseline and indicating that the RGB branch is effectively unused.

  Together these variants isolate modality dropout as the component responsible
  for the gains, with cross-attention contributing only when paired with dropout.

\section{Additional Quantitative Results}
\label{app:quant}

  \begin{table}[!t]
      \centering
      \resizebox{\textwidth}{!}{%
      \begin{tabular}{l|ccccccc}
      \hline
      Level & \textbf{DIPOLE} & RGB\_resnet18 & RGBD\_resnet18 & RGBD\_ViT & RGBD\_MultiMAE & PointCloud\_DP3 & PointCloud\_PonderV2 \\
      \hline
      L0 & \cellcolor{lightblue}\textbf{0.68} & 0.46 & 0.43 & 0.44 & 0.48 & 0.12 & 0.19 \\
      L1 & \cellcolor{lightblue}\textbf{0.64} & 0.17 & 0.23 & 0.36 & 0.32 & 0.13 & 0.15 \\
      L2 & \cellcolor{lightblue}\textbf{0.64} & 0.19 & 0.21 & 0.20 & 0.20 & 0.08 & 0.08 \\
      \hline
      \end{tabular}}
      \caption{{\bf Cross-randomization-level performance.} Each entry reports the average success rate of an encoder at a given randomization level, averaged over all tasks.}
      \label{tab:cross_randomization}
  \end{table}

  \begin{table}[h]
      \centering
      \resizebox{\textwidth}{!}{%
      \begin{tabular}{l|ccccccc}
      \hline
      Relative Dispersion & \textbf{DIPOLE} & RGB\_resnet18 & RGBD\_resnet18 & RGBD\_ViT & RGBD\_MultiMAE & PointCloud\_DP3 & PointCloud\_PonderV2 \\
      \hline
      Cross-task $\downarrow$ & \cellcolor{lightblue}\textbf{30\%} & 52.80\% & 48.48\% & 45.49\% & 43.86\% & 126.19\% & 133.33\% \\
      Cross-randomization-level $\downarrow$ & \cellcolor{lightblue}\textbf{2.53\%} & 32.72\% & 35.01\% & 45.28\% & 37.23\% & 5.06\% & 15.43\% \\
      \hline
      \end{tabular}}
      \caption{{\bf Cross-task and cross-randomization stability comparison.} We evaluate stability with relative dispersion of success rates across tasks or randomization levels. Lower relative fluctuation means more stable performance across different tasks or observation distributions.}
      \label{tab:stability}
  \end{table}

  \begin{table}[h]
    \centering
    \caption{Emergent capability of complementarity-aware fusion. On settings where both constituent unimodal baselines fail ($\leq$1\%) when trained alone, DIPOLE recovers to 72.3\% mean success.}
    \label{tab:emergence}
    \begin{tabular}{l|ccc}
        \hline
        Setting & ResNet (RGB only) & DP3 (PC only) & \textbf{DIPOLE} \\
        \hline
        AlphabetSoup L1 & 0.00 & 0.00 & \textbf{0.67} \\
        AlphabetSoup L2 & 0.00 & 0.00 & \textbf{0.73} \\
        Tomato L1 & 0.01 & 0.00 & \textbf{0.68} \\
        Tomato L2 & 0.01 & 0.00 & \textbf{0.81} \\
        \hline
        Mean & 0.005 & 0.00 & \textbf{0.723} \\
        \hline
    \end{tabular}
  \end{table}

\end{document}